\documentclass[letterpaper, 10 pt, conference]{IEEEtran}                            

\IEEEoverridecommandlockouts   

\title{\LARGE \bf
Carrying the uncarriable: a deformation-agnostic and human-\\cooperative framework for unwieldy objects using multiple robots}

\author{Doganay Sirintuna$^{1,2}$, Idil Ozdamar$^{1,2}$, and Arash Ajoudani$^{1}$
\thanks{$^{1}$ Human-Robot Interfaces and Interaction Laboratory, Istituto Italiano di Tecnologia, Genoa, Italy. {\tt \{doganay.sirintuna, idil.ozdamar, arash.ajoudani\}@iit.it}}
\thanks{$^{2}$ Dept. of Informatics, Bioengineering, Robotics, and System Engineering. University of Genoa, Genoa, Italy.}
\thanks{This work was supported in part by the ERC-StG Ergo-Lean (Grant Agreement No.850932), in part by the European Union’s Horizon 2020 research and innovation programme SOPHIA (Grant Agreement No. 871237).}
\thanks{The authors thank Juan M. Gandarias, Alberto Giammarino and Pietro Balatti for their contribution to the experimental setup and the fruitful discussions.}
}

\usepackage{cite}
\usepackage{amsmath}
\usepackage{amssymb}
\usepackage[pdftex]{graphicx}
\usepackage[dvipsnames]{xcolor}
\usepackage[hyphens]{url}
\usepackage{caption}
\usepackage{algorithm}
\usepackage{algpseudocode}
\usepackage[english]{datetime2}
\usepackage{tabularx}
\newcolumntype{P}[1]{>{\centering\arraybackslash}p{#1}}
\newcolumntype{M}[1]{>{\centering\arraybackslash}m{#1}}

\DeclareCaptionFont{mysize}{\fontsize{8}{9.6}\selectfont}
\begin{document}

\maketitle

\begin{abstract}
This manuscript introduces an object deformability-agnostic framework for co-carrying tasks that are shared between a person and multiple robots. Our approach allows the full control of the co-carrying trajectories by the person while sharing the load with multiple robots depending on the size and the weight of the object. This is achieved by merging the haptic information transferred through the object and the human motion information obtained from a motion capture system. One important advantage of the framework is that no strict internal communication is required between the robots, regardless of the object size and deformation characteristics. We validate the framework with two challenging real-world scenarios: co-transportation of a wooden rigid closet and a bulky box on top of forklift moving straps, with the latter characterizing deformable objects. In order to evaluate the generalizability of the proposed framework, a heterogenous team of two mobile manipulators that consist of an Omni-directional mobile base and a collaborative robotic arm with different DoFs is chosen for the experiments. The qualitative comparison between our controller and the baseline controller (i.e., an admittance controller) during these experiments demonstrated the effectiveness of the proposed framework especially when co-carrying deformable objects. Furthermore, we believe that the performance of our framework during the experiment with the lifting straps offers a promising solution for the co-transportation of bulky and ungraspable objects.
\end{abstract}

\section{Introduction}
\label{sec:introduction}

Transportation of objects is a commonly performed task in today's manufacturing environments such as factories and warehouses. This physically challenging operation for the workers often requires the collaboration of multiple partners. Considering the dynamically changing environments of industrial settings, it is difficult to have a solution where the robots cooperate with each other to execute the task autonomously. On the contrary, the human-robot teams that exploit human adaptability and decision-making skills can be employed to perform the task with the required flexibility in unstructured environments while reducing the ergonomic discomforts of workers \cite{ajoudani2018progress}.

Although the existence of a human partner brings significant contributions to the team, there are still challenges for possessing the required versatility for co-carrying various objects. First of all, an appropriate number of robot partners must be included in the operation according to the load and size of the object, since there is a maximum payload that a robot partner can handle. Moreover, the developed human-multi-robot system should effectively map the human motion intention to each robot. Despite the relatively large number of studies on this subject in the literature, most existing techniques require the object being transported to be rigid. As a result, the conventional haptic-based techniques may fall short if an object transmits incomplete wrenches due to its deformability.
\begin{figure}\vspace{-3mm}
    \centering
    \resizebox{0.70\columnwidth}{!}{\rotatebox{0}{\includegraphics[trim=10cm 2cm 12.5cm 5.1cm, clip=true]{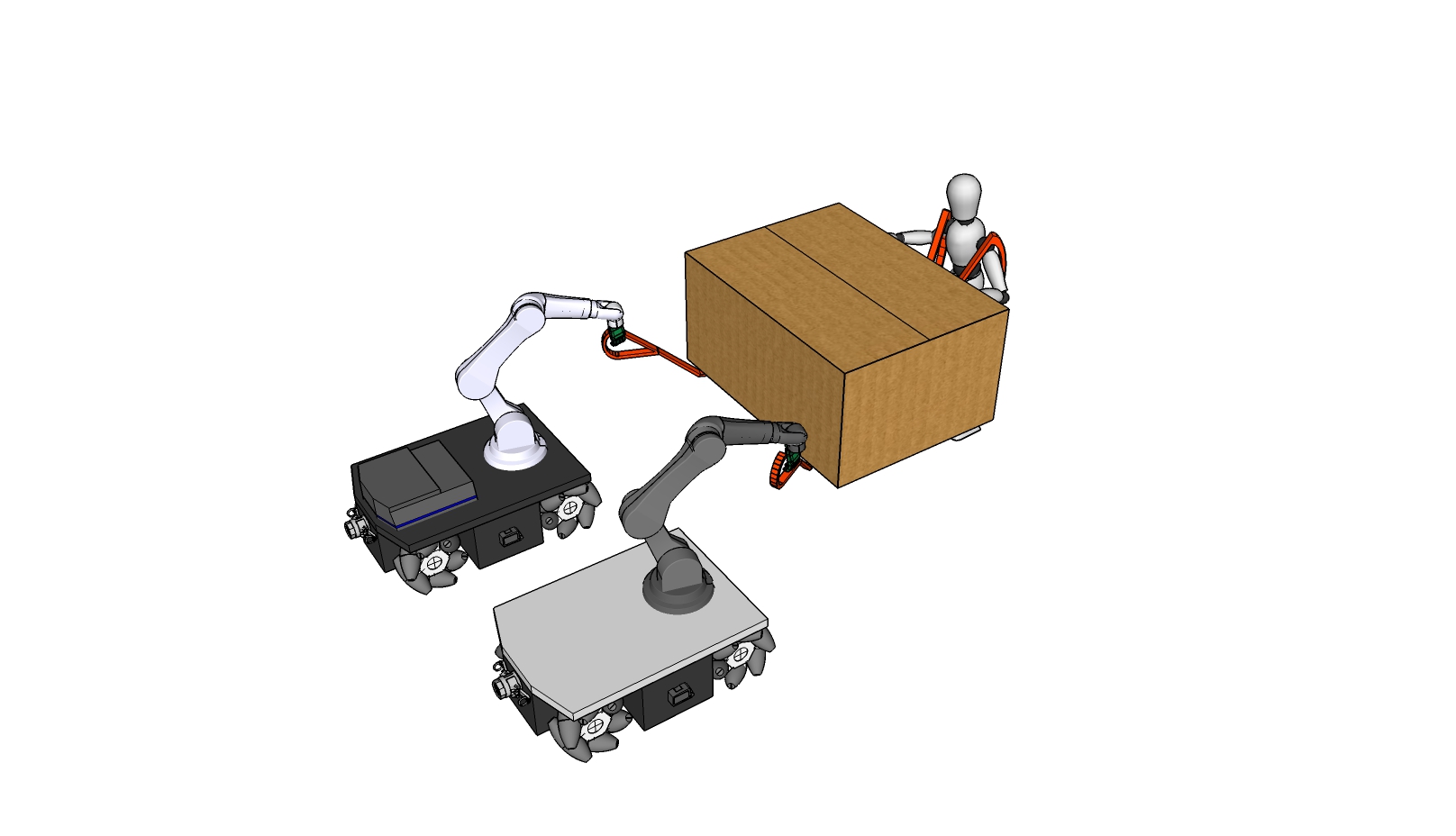}}}\vspace{-1.6mm}
    \caption{We propose an adaptive human-multi-robot framework to co-transport objects irrespective of their deformation characteristics.}
    \vspace{-0.8cm}
    \label{fig:digest_figure}
\end{figure}
To address these challenges, we present a human-multi-robot co-transportation framework where objects irrespective of their deformation characteristics can be handled. The proposed framework extends our earlier work~\cite{doganay2022} to adapt it to the scenarios where the object being carried is unwieldy (and maybe ungraspable), deformable, and impossible to manipulate with one single robot partner. Our approach allows the human operator to collaborate with any number of robotic partners when necessary, without demanding direct communication between them. Furthermore, since our framework does not rely on a particular structure of the robot partners, a heterogeneous team having robots with different Degrees of Freedom (DoFs) and controller types can be employed according to the task needs.

\begin{figure*}[!ht]
	\centering
	\includegraphics[width=0.88\linewidth]{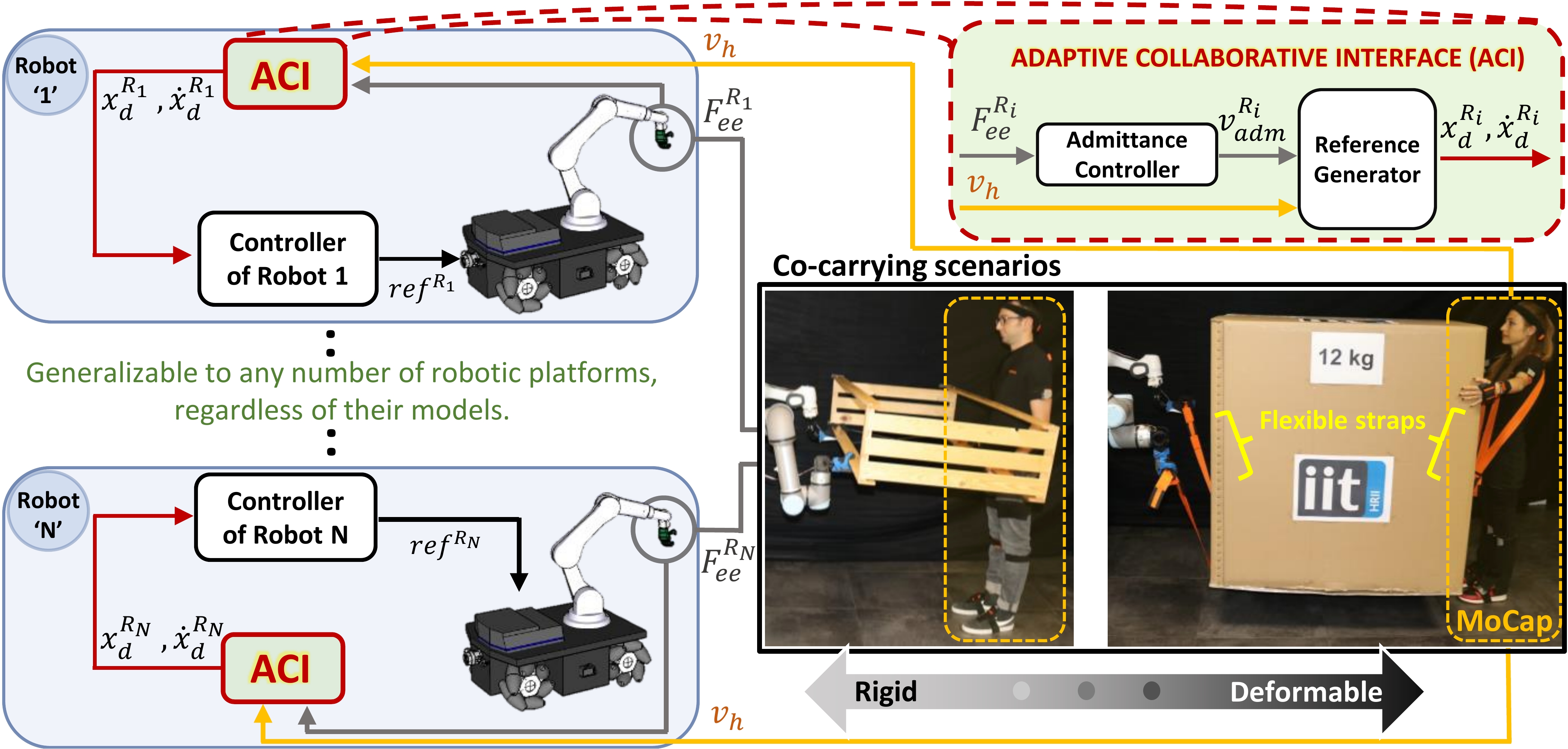}\vspace{-1mm}
	\caption{High-level scheme of the proposed control framework. The interaction forces at the end-effector, $F_{ee}$, and the hand velocity, $v_{h}$, are sent to the Adaptive Collaborative Interface (ACI) to generate the reference pose and twist exclusively for each robot.} 
    \vspace{-0.65cm}
	\label{fig:ACI}
\end{figure*}\vspace{-0mm}

The rest of this article is structured as follows: Section~\ref{sec:related_work} outlines the literature about collaborative carrying. In Section~\ref{sec:ACI}, we explain our adaptive framework that generates motion references for the robot partners by merging the haptic and human movement information. The rest of Section~\ref{sec:methodology} reports the details of our human-multi-robot system. Then, in Section~\ref{sec:experiments}, the conducted experiments to validate our framework by comparing its performance during co-transportation of objects having different deformation characteristics (e.g., a bulky box on forklift moving straps and a wooden closet) are introduced. Section~\ref{sec:results_and_discussion} includes the quantitative results of the experiments and the discussion of the overall framework performance. Finally, Section~\ref{sec:conclusion} draws the conclusions of the study.

\section{Related Work}
\label{sec:related_work}

In the physical human-robot interaction literature (pHRI), considerable attention has been given to the co-manipulation of jointly-held objects. In this topic, there exist multiple studies where the haptic information transmitted through the rigid object is employed in order to allow co-transportation by regulating the interaction between the human and the robot~\cite{Ryojun,Karayiannidis,Duchaine,duchaine2009safe,bussy2012proactive}.

However, very few studies confront the complexity of deformable object manipulation where only the deficient haptic information is available on the robot-end~\cite{sanchez2018robotic}. Maeda et al. propose a hybrid impedance controller based on the haptic and visual feedback acquired from the human operator~\cite{Maeda}. It utilizes the minimum jerk model of human hand motion to set the desired position of the robot. Although the proposed controller is validated during a one-dimensional point-to-point co-transportation of a deformable rubber pipe, a recent study~\cite{jerk_model} reveals that the minimum jerk model is not a proper fit for complex collaborative tasks. Similarly, a novel controller that merges the wrench information transferred through the object and the visual feedback is presented in~\cite{kruse2015collaborative} for co-manipulation of cloths. Nonetheless, this approach can be employed only for cloth-type objects, since it exploits the obtained knowledge from the detected deformed areas of the fabric to restore its tautness. In another framework that is less dependent on the physical state of the object~\cite{delpreto2019sharing}, DelPreto and Rus utilize electromyography (EMG) signals acquired from the upper arm of the operator to perform collaborative lifting tasks. However, this framework remains limited only to the co-lifting scenarios with 1-D motion.

Considering the studies mentioned above, the cooperation of a human with a single robot is a well-studied research topic. However, to the best of the authors’ knowledge, not much prior work has been done on human-multi-robot co-transportation, and especially when it comes to the deformation ranges of the target objects. In~\cite{WangNrobot}, the force applied by the leader (human or robot) is amplified by the other robots in the team, in order to provide the required assistance during the co-manipulation of the jointly grasped object. Furthermore, their proposed approach is a scalable and easily configurable solution for co-transportation without requiring communication among the robots. Recently, Matthew et al.~\cite{omnibot} built the Omnid mobile collaborative robot, named 'mocobot', designed specifically to facilitate human-robot teaming for object transportation. They validate that the mocobot team under the guidance of a human during manipulation of a large solid PVC pipe assembly and an articulated payload. Nonetheless, these studies still require a rigid human-robot connection while the employed robots are identical. On the other hand, frameworks that are able to have a heterogeneous group of robots can exploit the diverse capabilities of the team to perform better, especially in complex scenarios~\cite{TuciReview}.

\section{Multi-Robot Co-Carry System Overview}
\label{sec:methodology}
The proposed multi-robot co-carrying framework consists of an Adaptive Collaborative Interface (ACI) that generates reference end-effector poses and twists for each robot, a decentralized control unit formed by robot loco-manipulation controllers, and a motion capture system (MoCap) to measure human movements. In this work, we chose two mobile manipulators one with a torque-controlled 7-DoF arm and the other with a position-controlled 6-DoF arm to demonstrate the scalability of the framework.
\vspace{-1mm}
\subsection{Adaptive Collaborative Interface}
\label{sec:ACI}
Adaptive Collaborative Interface generates exclusive movement references for the robots to co-manipulate objects having different deformation characteristics (see Fig.~\ref{fig:ACI}). Each robot in the framework will obtain its desired end-effector motion via an ACI that uses conveyed force through the object and human hand velocity as inputs. The interface consists of two operational sub-units, namely an \emph{Admittance Controller} which calculates a reference velocity from the force input, and a \emph{Reference Generator} that computes desired reference pose and twist of the robot by merging the velocity calculated by the Admittance Controller and the hand velocity acquired from the MoCap system.
\subsubsection{Admittance Controller}
The admittance controllers of each robot can be expressed in the Laplace domain as:  
\begin{equation}
  \boldsymbol{{V}}_{adm}^{R_{i}}(s) = \frac{\boldsymbol{F}_H^{R_{i}}(s)}{\boldsymbol{M}_{adm}^{R_{i}}s+\boldsymbol{D}_{adm}^{R_{i}}},
\end{equation}
\noindent where $\boldsymbol{V}_{adm}^{R_{i}}(s)\in\mathbb{R}^{3}$ is the Laplace transform of the admittance reference translational velocity, $\boldsymbol{F}_H^{R_{i}}(s)\in\mathbb{R}^{3}$ is the Laplace transform of the measured forces, $\boldsymbol{M}_{adm}^{R_{i}}$, $\boldsymbol{D}_{adm}^{R_{i}}$ $\in$ $\mathbb{R}^{3\times3}$ are the desired mass and damping matrices, $s$ is the Laplace variable, and \textit{i} is used to represent the robot id.
\subsubsection{Reference Generator}
During the co-transportation of deformable objects, relying solely on haptic information does not guarantee effective coordination between human and robot partners. In order to address this issue, we introduced the \textit{Reference Generator} unit which calculates the desired motion (pose and twist), $\boldsymbol{{x}}_{d}^{R_{i}}$ and $\boldsymbol{\dot{x}}_{d}^{R_{i}}$, of each robot in the team by combining its admittance reference velocity, $v_{adm}^{R_{i}}(t)$, and the human hand velocity, $v_{h}(t)$, through an adaptive index, $\alpha^{R_{i}}(t)$. This index is formulated in the following way to infer the deformability of the object being carried and leads us to properly regulate $v_{adm}^{R_{i}}(t)$ and $v_{h}(t)$ contributions within the desired end-effector velocity, $v_{d}^{R_{i}}(t)$:
\begin{equation}
    \alpha^{R_{i}}(t) = 1-\frac{\int_{t_{c}-W_{l}}^{t_{c}} ||\boldsymbol{{v}}_{adm}^{R_{i}}(t)|| \,dt\ }{\int_{t_{c}-W_{l}}^{t_{c}} ||\boldsymbol{{v}}_{h}(t)|| \,dt\  + \epsilon};
\label{eq:alfa}
\end{equation}
\begin{equation}
    \boldsymbol{{v}}_{d}^{R_{i}}(t) = \boldsymbol{{v}}_{adm}^{R_{i}}(t) + \alpha^{R_{i}}(t)\boldsymbol{{v}}_{h}(t), 
\label{eq:vd}
\end{equation}
\noindent where $\alpha^{R_{i}}(t) \in [0,1]$ is the adaptive index of the \textit{i}-th robot, which is saturated at 0, $t_c$ is the current time, $W_l$ is the length of the sliding time window and $\epsilon$ is a small number to prevent division by zero problem. Hence, this index allows us to normalize the object between non-deformable ($\alpha(t) = 0$) and highly deformable ($\alpha(t) = 1$). In order to further clarify how the \textit{Reference Generator} operates to deal with objects having different deformation properties, three cases can be defined as presented in Table~\ref{table:deformability_table} (see \cite{doganay2022} for more details).
\begin{table}[!b]\vspace{-5mm}
\centering
\caption{Object Deformability Cases}\vspace{-2mm}
\label{table:deformability_table}
\begin{tabular}{|M{2.05cm}|M{1.5cm}|M{1cm}|M{2.35cm}| }
 \hline
 \hspace{-2mm}\textbf{Object} & \hspace{-2mm}$\boldsymbol{{v}}_{adm}$ (Eq. 1) & \hspace{-2mm}$\boldsymbol{\alpha}$ (Eq. 2) & \hspace{-2mm}$\boldsymbol{{v}}_{d}$ (Eq. 3) \\
 \hline
 \hspace{-2mm}Highly Deformable   (e.g., loose rope)  & \hspace{-2mm}$\approx 0$  & \hspace{-2mm}$\approx 1$ & \hspace{-2mm}$\approx \boldsymbol{{v}}_{h}(t)$  \\
 \hline
 \hspace{-2mm}Non-deformable (e.g., rigid rod)  & \hspace{-2mm}$\approx \boldsymbol{{v}}_{h}(t) $ & \hspace{-2mm}$\approx 0$ & \hspace{-2mm}$\approx \boldsymbol{{v}}_{adm(t)} \approx \boldsymbol{{v}}_{h}(t)$  \\
 \hline
Partially Deformable  & \hspace{-2mm}$v_{adm}(t)$  & \hspace{-2mm}$\alpha(t)$  & \hspace{-1mm}$v_{adm}(t) + \alpha(t)v_{h}(t)$  \\
 \hline
\end{tabular}\vspace{-0mm}
\end{table}

Next, the desired pose and twist are sent to the whole-body loco-manipulation controller of each robot calculated by $\boldsymbol{{x}}_{d}^{R_{i}}(t) = \int_{0}^{t} \boldsymbol{\dot{x}}_{d}^{R_{i}}(t) \,dt $, where $\boldsymbol{\dot{x}}_{d}^{R_{i}}(t) = [{{v}_{d}^{R_{i}}(t)}^{T}, {0}^{T}]^T$.

\subsection{The Robotic Platforms}\label{subsec:robotic_platform}
\subsubsection{The MObile Collaborative robotic Assistant (MOCA) Platform}
\begin{figure*}[!ht]\vspace{-0mm}
	\centering
	\includegraphics[width=1\linewidth]{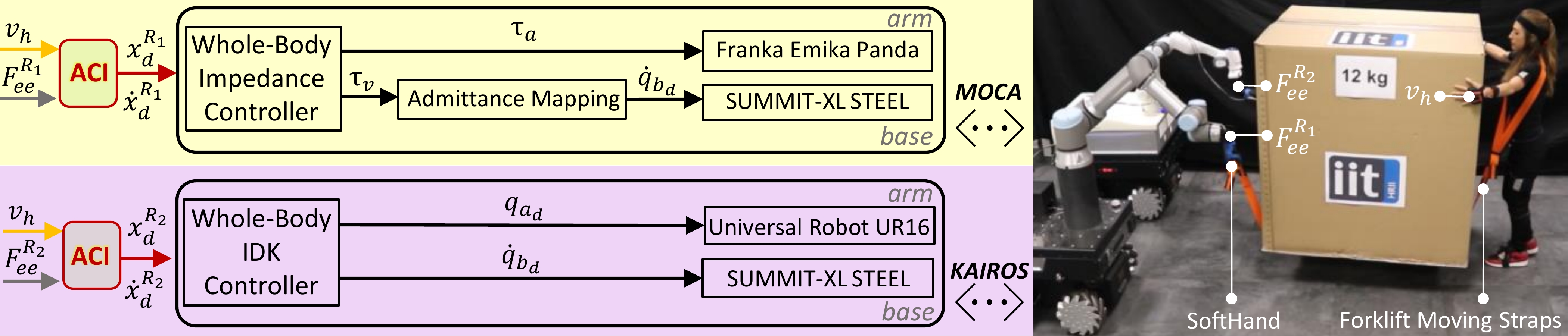}\vspace{-1.5mm}
	\caption{The detailed block diagrams refer to the whole-body controllers of the MOCA (yellow colored) and the KAIROS (purple colored) robotic platforms. 
	} 
    \vspace{-0.65cm}
	\label{fig:controllers_diagram}
\end{figure*}

\paragraph{Hardware Details}
The MOCA (see Fig.~\ref{fig:controllers_diagram}) comprises an Omni-directional Robotnik SUMMIT-XL STEEL mobile platform and a lightweight torque-controlled 7 DoFs Franka Emika Panda as a manipulator \cite{wu2019teleoperation}. 
Besides, an underactuated Pisa/IIT SoftHand was mounted at the end-effector of the arm to grasp the object being carried. For the precise external force measurements, an additional F/T sensor is placed between the robot’s flange and the SoftHand. 
\paragraph{Weighted Whole-Body Cartesian Impedance Controller} 
The formulation of the whole-body decoupled dynamics of the MOCA, where the comprehensive analysis can be found in \cite{wu2021unified}, is as follows:
\begin{gather}
\begin{aligned}
\label{eq:whole_body_dynamics}
&\overbrace{\begin{pmatrix}
 \boldsymbol{M}_b & \boldsymbol{0} \\
 \boldsymbol{0} & \boldsymbol{M}_a(\boldsymbol{q}_a)
\end{pmatrix}}^{\boldsymbol{M}}
\overbrace{\begin{pmatrix}
 \boldsymbol{\ddot{q}}_{b} \\
 \boldsymbol{\ddot{q}}_{a}
\end{pmatrix}}^{\boldsymbol{\ddot{q}}} +
\overbrace{\begin{pmatrix}
 \boldsymbol{D}_b & \boldsymbol{0} \\
 \boldsymbol{0} & \boldsymbol{C}_a(\boldsymbol{q}_a,\dot{\boldsymbol{q}}_a)
\end{pmatrix}}^{\boldsymbol{C}}
\overbrace{\begin{pmatrix}
 \boldsymbol{\dot{q}}_{b} \\
 \boldsymbol{\dot{q}}_{a}
\end{pmatrix}}^{\boldsymbol{\dot{q}}} + \\
&\underbrace{\begin{pmatrix}
 \boldsymbol{0} \\
 \boldsymbol{g}_a(\boldsymbol{q}_a)
\end{pmatrix}}_{\boldsymbol{g}} = 
\underbrace{\begin{pmatrix}
 \boldsymbol{\tau}_{b} \\
 \boldsymbol{\tau}_{a}
\end{pmatrix}}_{\boldsymbol{\tau}_{c}} + 
\underbrace{\begin{pmatrix}
 \boldsymbol{\tau}_{b,ext} \\
 \boldsymbol{\tau}_{a,ext}
\end{pmatrix}}_{\boldsymbol{\tau}_{ext}},
\raisetag{30pt}
\end{aligned}
\end{gather}
where $\boldsymbol{M}_{b}$  $\in$ $\mathbb{R}^{n_{b}\times n_{b}}$ and $\boldsymbol{D}_{b}$  $\in$ $\mathbb{R}^{n_{b}\times n_{b}}$ are the diagonal positive definite virtual inertia and virtual damping of the mobile base,  $\boldsymbol{\dot{q}}_{b}$, $\boldsymbol{\ddot{q}}_{b}$ $\in$ $\mathbb{R}^{n_{b}}$
are its input velocity and the acceleration, $\boldsymbol{\tau}_{b}$ $\in$ $\mathbb{R}^{n_{b}}$ and $\boldsymbol{\tau}_{b,ext}$ $\in$ $\mathbb{R}^{n_{b}}$ are the virtual and external torques being $n_{b}$ is the DoF of the base. Regarding the arm, 
$\boldsymbol{M}_{a}$  $\in$ $\mathbb{R}^{n_{a}\times n_{a}}$
is the symmetric and
positive definite inertia matrix,  $\boldsymbol{C}_{a}$ $\in$ $\mathbb{R}^{n_{a}}$ is the Coriolis and centrifugal force, $\boldsymbol{{q}}_{a}$, $\boldsymbol{\dot{q}}_{a}$, $\boldsymbol{\ddot{q}}_{a}$ $\in$ $\mathbb{R}^{n_{a}}$ are the joint angles, velocities and accelerations vectors, $\boldsymbol{g}_{a}$ $\in$ $\mathbb{R}^{n_{a}}$ is the gravity vector, $\boldsymbol{\tau}_{a}$ $\in$ $\mathbb{R}^{n_{a}}$ and $\boldsymbol{\tau}_{a,ext}$ $\in$ $\mathbb{R}^{n_{a}}$ are the commanded and the external torques being $n_{a}$ is the DoF of the arm. Lastly, $\boldsymbol{\tau}_{c}$ $\in$ $\mathbb{R}^{n_{a}+n_{b}}$ and $\boldsymbol{\tau}_{ext}$ $\in$ $\mathbb{R}^{n_{a}+n_{b}}$ denote joint-space input and external torques.
The whole-body Cartesian impedance controller of MOCA computes the reference joint torques,${\boldsymbol{\tau}_c}$, according to the solution of the prioritized weighted inverse dynamics algorithm. The optimization problem of finding the closest ${\boldsymbol{\tau}_c}$ to some desired ${\boldsymbol{\tau}_0}$ that realizes the operational forces, $\boldsymbol{F}$, can be written as:
\begin{equation}
\label{eq:secondary_task}
    \min_{\boldsymbol{\tau}_c}\frac{1}{2}||\boldsymbol{\tau}_c - \boldsymbol{\tau}_0||^2_{\boldsymbol{W}}
     \qquad \textrm{s.t.}\: \boldsymbol{\bar{J}}^T\boldsymbol{\tau}_c  =  \boldsymbol{F},
\end{equation}
where ${\boldsymbol{W}} \in \mathbb{R}^{(n_{a}+n_{b})\times (n_{a}+n_{b})}$ is a positive definite weighting matrix, $\boldsymbol{\bar{J}}^T={\big(\boldsymbol{J}\boldsymbol{M}^{-1} \boldsymbol{J}^{T} \big)}^{-1}\boldsymbol{J}\boldsymbol{M}^{-1} $ is the dynamically consistent pseudo-inverse of $\boldsymbol{J(q)}$, and the constraint $\boldsymbol{\bar{J}}^T\boldsymbol{\tau}_c=\boldsymbol{F}$ is the general relationship between the generalised joint torques and the operational forces~\cite{KhatibMoca}.

The closed-form solution to this problem can be formulated as:
\begin{equation}\label{eq:opt_solution}
\begin{aligned}
    \boldsymbol{\tau}_c & = \boldsymbol{W^{-1}M^{-1}J^{T}\Lambda_{W}\Lambda^{-1}F} \\
    & + (\boldsymbol{I-W^{-1}M^{-1}J^{T}\Lambda_{W}JM^{-1}})\boldsymbol{\tau}_0 ,
\end{aligned}
\end{equation}
where $\boldsymbol{\Lambda}= {\big(\boldsymbol{J}\boldsymbol{M}^{-1}\boldsymbol{J}^{T}\big)}^{-1}$ is the Cartesian inertia and $\boldsymbol{\Lambda_{W}}=\boldsymbol{J}^{-T}\boldsymbol{MWM}\boldsymbol{J}^{-1}$ is the weighted Cartesian inertia, analogous to $\boldsymbol{\Lambda}$. The positive definite weighting matrix $\boldsymbol{W}$ is defined as $\boldsymbol{W}(\boldsymbol{q})=\boldsymbol{H}^T\boldsymbol{M}^{-1}(\boldsymbol{q})\boldsymbol{H}$, where ${\boldsymbol{H}} \in \mathbb{R}^{(n_{a}+n_{b})\times(n_{a}+n_{b})}$ is a diagonal matrix which can be tuned to produce different mobility modes of the base and arm~\cite{edoardo}.

To obtain the desired Cartesian impedance behavior, $\boldsymbol{F}$ is computed as $\boldsymbol{F} =   \boldsymbol{D}_d(\dot{\boldsymbol{x}}_d-\dot{\boldsymbol{x}}) + \boldsymbol{K}_d(\boldsymbol{x}_d - \boldsymbol{x})$, where $\boldsymbol{x}, \boldsymbol{x}_d \in \mathbb{R}^{6}$ and $\dot{\boldsymbol{x}}, \dot{\boldsymbol{x}_d} \in \mathbb{R}^{6}$ are the current and desired Cartesian poses and twists, and $\boldsymbol{D}_d,\boldsymbol{K}_d \in \mathbb{R}^{6 \times 6}$ are the desired Cartesian damping and stiffness matrices, respectively. 

Lastly, the desired null-space behavior of the robot is obtained by calculating $\boldsymbol{\tau}_0$ as $\boldsymbol{\tau}_{0} = -\boldsymbol{D}_0\dot{\boldsymbol{q}} - \boldsymbol{K}_0(\boldsymbol{q}- \boldsymbol{q}_{0})$, where $\boldsymbol{q}_{0}$ is the desired joint configuration, ${\boldsymbol{K}_0}$, ${\boldsymbol{D}_0} \in \mathbb{R}^{(n_{a}+n_{b})\times(n_{a}+n_{b})}$ represent the desired joint-space stiffness and damping.

\subsubsection{The Kairos Platform}
\paragraph{Hardware Details}
The Kairos (see Fig.~\ref{fig:controllers_diagram}) consists of an Omni-directional Robotnik SUMMIT-XL STEEL mobile base, and a high-payload (16 kg) 6-DoFs Universal Robot UR16e arm attached on top of the base and equipped with a Pisa/IIT Softhand. This manipulator also contains an F/T sensor that measures the applied wrenches at the robot’s flange.

\paragraph{Weighted Whole-body Closed-Loop Inverse Kinematics Controller}

The whole-body controller utilized on the Kairos computes the desired joint velocities $\boldsymbol{\dot{q}}_d$ $\in$ $\mathbb{R}^{n_{a}+n_{b}}$ by solving a Hierarchical Quadratic Program (HQP) problem composed of two tasks, where $n_{a}$ and $n_{b}$ denote the DoF of the base and the arm. This formulation of the problem enables us to exploit the redundancy of the robotic platform while distributing the movements between the mobile base and the arm. The higher priority task is written as the following cost function (dependencies are dropped)~\cite{seraji1990improved}, in order to track the desired end-effector motion ($\boldsymbol{\dot{x}}_d$ and ${\boldsymbol{x}_d}$):
\begin{align}
    \mathcal{L}_1 = || \dot{\boldsymbol{x}_d} + \boldsymbol{K}({\boldsymbol{x}_d} -{\boldsymbol{x}}) - \boldsymbol{J}\dot{\boldsymbol{q}}||^2_{\boldsymbol{W}_{1}} + ||k\dot{\boldsymbol{q}}||^2_{\boldsymbol{W}_{2}},
\end{align}
where $\boldsymbol{\dot{q}}$ $\in$ $\mathbb{R}^{n_{a}+n_{b}}$ is the optimization variable, ${\boldsymbol{J}}$ $\in$ $\mathbb{R}^{6\times(n_{a}+n_{b})}$ is the whole-body Jacobian, ${\boldsymbol{x}}$ $\in$ $\mathbb{R}^{6}$ is the current end-effector pose, ${\boldsymbol{W}_1}$ $\in$ $\mathbb{R}^{6\times6}$, ${\boldsymbol{W}_2}$ $\in$ $\mathbb{R}^{(n_{a}+n_{b})\times(n_{a}+n_{b})}$, and ${\boldsymbol{K}}$ $\in$ $\mathbb{R}^{6\times6}$ are diagonal positive definite matrices and $k$ $\in$ $\mathbb{R}_{>0}$ is the so-called damping factor \cite{chiaverini1992weighted}, which changes according to the manipulability index of the arm \cite{deo1995overview, hollerbach1987redundancy, wampler1986manipulator}.

The cost function of the secondary task which keeps the arm close to the default joint configuration, $\boldsymbol{q}_{0}$, can be formulated as~\cite{nakanishi2005comparative} $  \mathcal{L}_2 = ||\boldsymbol{q}_{0} - \boldsymbol{q}||^2_{\boldsymbol{W}_{3}}$, where ${\boldsymbol{W}_3}$ $\in$ $\mathbb{R}^{(n_{a}+n_{b})\times(n_{a}+n_{b})}$ is a diagonal positive semidefinite matrix. The desired joint velocities that minimize this cost are calculated as the negative gradient of it. Later, these velocities are projected onto the null-space of the first task.

\subsection{MoCap System}\label{subsec:mocap}

To track human movements, we decided to use the Xsens as a MoCap system (see Fig.~\ref{fig:controllers_diagram}) due to its precision and robustness. It is composed of seventeen Inertial Measurement Units (IMUs) distributed on the human body. 
This system allows us to obtain real-time measurements of the human hand which are used as an input for our controller (see Section~\ref{sec:ACI}).
Note that, our framework is able to operate even though Xsens is replaced with alternative MoCap systems, such as vision-based ones.

\section{EXPERIMENTS}
\label{sec:experiments}

In this section, we first describe the two real-world collaborative transportation scenarios designed to evaluate the performance of our proposed framework by comparing it with a baseline controller (i.e., an admittance controller). Then, the selected controller parameters for the experiments are reported for both robots. The video accompanying this paper, also available at \url{https://youtu.be/Q3sA6YzTaaE}, includes the demonstrations with our controller and the baseline controller.

\subsection{Experimental Scenarios}
\label{subsec:experimental_scenario}

In the experiments, the subjects were asked to co-carry 1) \textit{a Bulky Box with Forklift Moving Straps} and 2) \textit{a Rigid Closet} along a designed path while collaborating with the robot team. The objects were intentionally chosen as oversized and heavy since they possess more challenging transportation problems than their smaller alternatives. The path comprises three sequential sub-movements which are backwards ($\approx$ 120 cm), sideways ($\approx$ 80 cm), and down-up ($\approx$ 20 cm both) with respect to the human. Both scenarios were executed once for each controller by the participants (1 female and 1 male, 25 years old) and the repetitions were not confined within a time limit. The whole experimental procedure was in accordance with the Declaration of Helsinki, and the protocol was approved by the ethics committee Azienda Sanitaria Locale (ASL) Genovese N.3 (Protocol IIT\_HRII\_ERGOLEAN 156/2020). 

\begin{figure*}\vspace{0mm}
    \centering
    \resizebox{0.8\textwidth}{!}{\rotatebox{0}{\includegraphics[trim=0cm 0cm 0cm 0cm, clip=true]{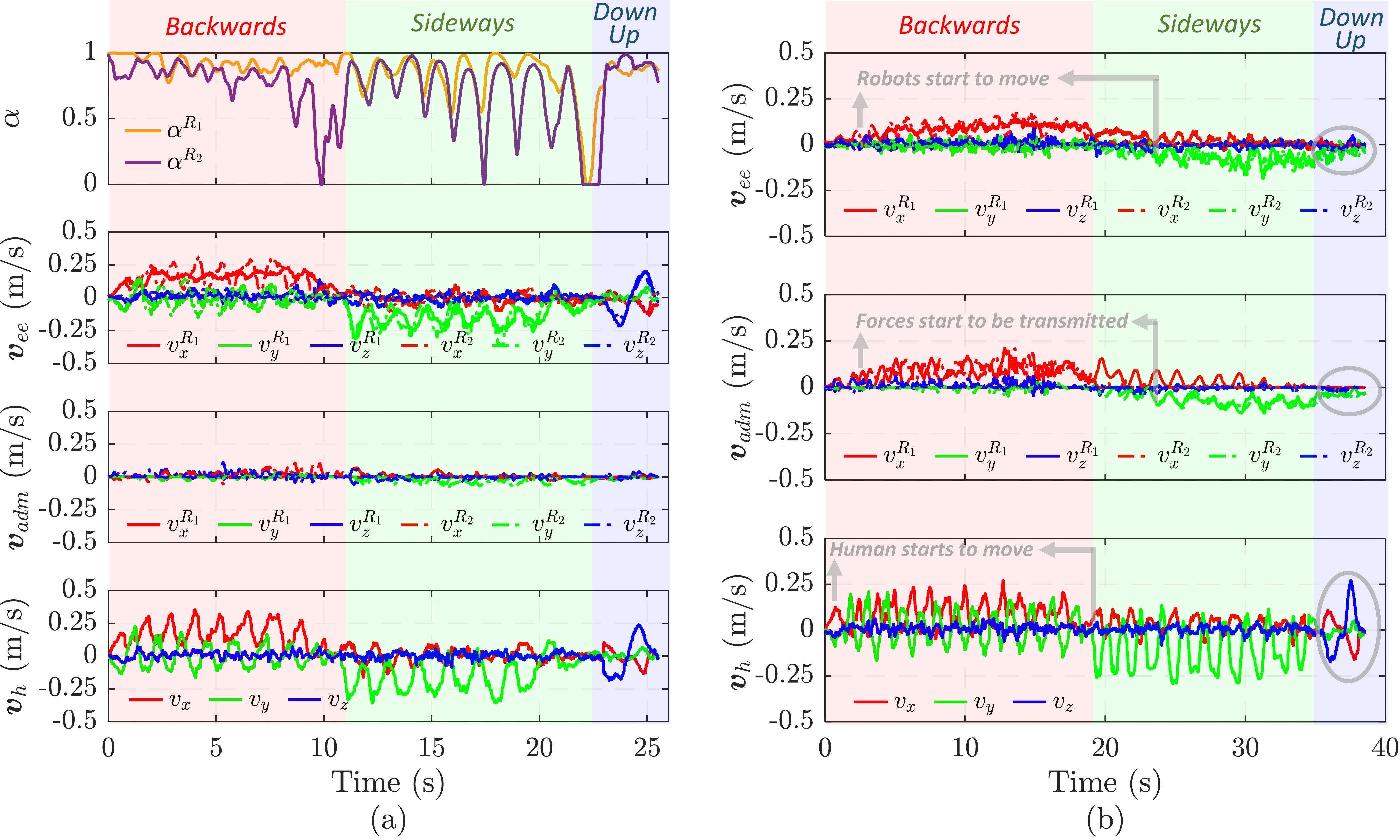}}}\vspace{-2mm}
     \caption{The ACI (a) and the admittance controller (b) results with regard to the co-transportation of the bulky box carried by forklift moving straps. The top left graph demonstrates the change in the adaptive index ($\alpha$) of robot 1 ($R_{1}$: yellow color) and robot 2 ($R_{2}$: purple color) throughout the sub-movements. The rest of the graphs show the end-effector velocity $v_{ee}$ and the admittance reference velocity $v_{adm}$ of robots ($R_{1}$: solid lines, $R_{2}$: dashed lines), the human hand velocity $v_{h}$.}
    \vspace{-0.7cm}
    \label{fig:experiment}
\end{figure*}  

\subsubsection{Co-carry of a Bulky Box with Forklift Moving Straps}
\label{subsubsec:scenario_1}

In this scenario, a 12 kg box with dimensions of 110 $\times$ 90 $\times$ 120 cm on crisscross lifting straps was co-transported where the robots grasped the ends on one side, and the ones on the opposite side were hung to the shoulders of the human (see Fig.~\ref{fig:controllers_diagram}). 
These lifting straps were prevalent commercially available tools to facilitate human-human co-transportation of objects even if they are not suitable to grasp (e.g., a large box). However, they pose a major challenge for human-robot teams since the transmitted forces and moments to the robots do not follow a rigid transformation, and the overall object can be considered deformable and even varying in different movement directions. 

\subsubsection{Co-carry of a Rigid Closet}
\label{subsubsec:scenario_2}
For this experiment, a 6 kg wooden closet with dimensions of 80 $\times$ 30 $\times$ 170 cm was chosen. Similar to the previous scenario, two robots grasped the object directly from one side and the human held it from the opposite while facing the robots during the co-manipulation (see Fig.~\ref{fig:ACI}). On the other hand, in this scenario, the rigid connection between partners allowed the transmission of the force applied by the human through the object and consequently to the robots.

\subsection{Controller Parameters}
\label{subsec:controller_parameters}

In all experiments, we used the same desired mass ($\boldsymbol{M}_{adm}=diag\{4,4,4\}$) and damping ($\boldsymbol{D}_{adm}=diag\{45,45,45\}$) for both robots. With this choice, we are able to make a fair comparison between the controllers (the ACI and the admittance controller) for the two scenarios. The sliding time window length ($W_l$), which is used to compute the adaptive index ($\alpha$) through Eq.~\eqref{eq:alfa}, was set to 0.5 s.
We adjusted the value of the $W_l$ to accurately identify the deformability of the object without having a delay that would negatively impact the task performance.

\subsubsection{Whole-body Controller of the MOCA Platform}

We experimentally tuned the Cartesian stiffness and the damping values to have an accurate desired end-effector motion while avoiding the unstable behavior of the robot.  
Therefore, these values were set to $\boldsymbol{K}_d=diag\{200,200,200,30,30,30\}$, and $\boldsymbol{D}_d=2\xi \boldsymbol{K}_d^{\frac{1}{2}}$, with $\xi=0.7$.
In addition, the mass and the damping of the mobile base were selected as $\boldsymbol{M}_{b} = diag \{105,105,210\}$ and $\boldsymbol{D}_{b} = 10M_{b}$ in order to ensure jerky and slow movements were not allowed. 
For having an adequate locomotion behaviour with MOCA the ${\boldsymbol{H}}$, ${\boldsymbol{K_{0}}}$ and ${\boldsymbol{D_{0}}}$ were assigned to ${\boldsymbol{H}}=\boldsymbol{I}_{n_{a}+n_{b}}$, ${\boldsymbol{K_{0}}}=diag\{50\cdot\boldsymbol{1}_{n_{a}+n_{b}}\}$ with $n_a$ and $n_b$ being 7 and 3, respectively, and $\boldsymbol{D}_0=2\xi\boldsymbol{K}_0^{\frac{1}{2}}$ where $\xi$ was 0.7.

\subsubsection{Whole-body Controller of the Kairos Platform}

To choose suitable values for $\boldsymbol{K}$, $\boldsymbol{W}_1$ and $\boldsymbol{W}_2$, the accurate tracking of the desired motion while preventing too high joint velocities has been considered. For this purpose, these parameters were experimentally selected as $\boldsymbol{K}=diag\{0.1,0.1,0.1,0.01,0.01,0.01\}$, $\boldsymbol{W}_1=100\cdot diag\{10,10,10,5,5,5\}$ and $\boldsymbol{W}_2=diag\{\boldsymbol{10}_{n_b}, \boldsymbol{0.5}_{n_a}\}$ where $n_a=6$ and $n_b=3$. Besides, to guarantee a locomotion behaviour where the most movement is assigned to the base instead of the arm, $\boldsymbol{W}_3$ was set to $diag\{\boldsymbol{0}_{n_b},\boldsymbol{1}_{n_a}\}$.

\section{RESULTS AND DISCUSSION}
\label{sec:results_and_discussion}

The results of the first experimental scenario where a bulky box was co-transported using forklift moving straps with (a) the proposed controller and (b) the admittance controller are depicted in Fig.~\ref{fig:experiment}. The top graph of Fig.~\ref{fig:experiment}a demonstrates the change in the adaptive index $\alpha$ for both robots throughout the task. The remaining plots of both experiments show the end-effector velocity $v_{ee}$, and the admittance reference velocity $v_{adm}$ of the robots, and the human hand velocity $v_{h}$.

Fig.~\ref{fig:experiment}a shows that $v_{adm}$ was close to 0 for both robots in all sub-movements of the experiment, even though the human operator moves along the desired trajectory (see $v_{h}$). It explains that although the object being carried is a rigid box, the forklift straps may not be stretched enough in non-loaded directions, to transfer the forces applied by the human because of its deformability. In this scenario where the haptic feedback was not available for both robots, the proposed controller allowed following the human operator successfully by using the human hand kinematic information. Except for the small jumps where the $\alpha$ values decreased because $\boldsymbol{{v}}_{h} \approx \boldsymbol{{v}}_{adm} \approx 0$, the high values of independently calculated adaptive indices indicated the controllers of both robots utilize the $v_{h}$ actively throughout the experiment.

On the other hand, when the admittance controller was employed (see Fig.~\ref{fig:experiment}b), the movements of both robots could start only after the straps were pulled sufficiently to transmit the forces. This situation can be observed especially in the beginnings of backwards and sideways sub-movements of the experiment (see the grey arrows). However, during the lowering and lifting parts, both robots were not able to move because the straps were not stretched enough (see the grey circles).

\begin{figure}\vspace{-0mm}
    \centering
    \resizebox{0.87\columnwidth}{!}{\rotatebox{0}{\includegraphics[trim=0cm 0cm 0cm 0cm, clip=true]{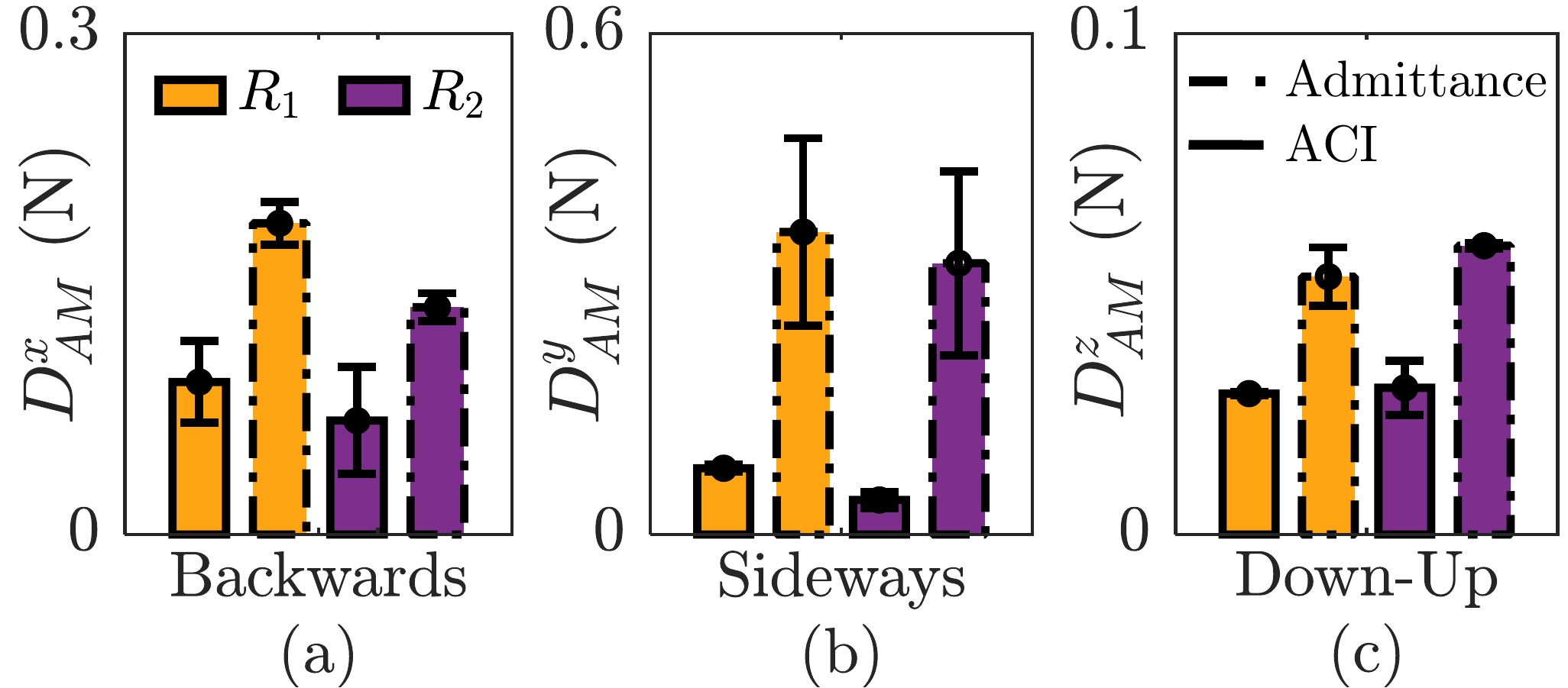}}}\vspace{-2mm}
     \caption{ The means and the standard errors of the alignment metric during sub-movements; backwards ($D^{x}_{AM}$) (a), sideways ($D^{y}_{AM}$) (b), and down-up ($D^{z}_{AM}$) (c) for the transportation of the bulky box with straps. The dashed and solid lines stand for the admittance controller and the ACI, and the yellow and purple colors represent the two robots used in the experiments.}
    \vspace{-0.7cm}
    \label{fig:Dam}
\end{figure}  

Moreover, we formulated an alignment metric to evaluate how accurately robots can follow the human partner despite the deformability of the straps. 
It is computed as: 
\begin{equation}
  \begin{aligned}
    {D_{AM}^*} &= \frac{\int_{t_{s}}^{t_{e}}{||\vec{R(t)^*} ||dt}}{t_{e}-t_{s}}; \\ 
    {\vec{R}(t)} = \boldsymbol{r}_{cee}(t) &- \boldsymbol{r}_{chh}(t) - (\boldsymbol{r}_{see}- \boldsymbol{r}_{shh}),
  \end{aligned} 
\label{eq:DAM}
\end{equation}
where $\vec{R(t)}$ is the difference vector between initial and current alignments of the human relative to the end-effector of the robot with * being X, Y, and Z components of the $\vec{R(t)}$ and $D_{AM}$. The current end-effector and current human hand positions ($\boldsymbol{r}_{cee}$ and $\boldsymbol{r}_{chh}$) and the starting end-effector and starting human hand positions ($\boldsymbol{r}_{see}$ and $\boldsymbol{r}_{shh}$) are used to calculate $\vec{R(t)}$. The start and the end time of the experiment are indicated by $t_s$ and $t_e$, respectively. If a robot can always preserve its initial configuration (i.e., ideal alignment) with respect to the human during the task execution $\vec{R(t)}$ will be zero. Consequently, $D_{AM}$ will be 0 too. However, all deviations from the ideal alignment will lead increase of $D_{AM}$.

Fig.~\ref{fig:Dam} reports the means and the standard errors of the alignment metric in the direction of the sub-movements (backwards; $D^{x}_{AM}$, sideways; $D^{y}_{AM}$, and Down-Up: $D^{z}_{AM}$) for the first experimental scenario.
From the results, we can deduce that when ACI was employed, both robots were better aligned with the participants in all sub-movements seeing $D^{x}_{AM}$, $D^{y}_{AM}$ and $D^{z}_{AM}$ values were higher for the admittance controller compared to ACI. 
Note that, since the distances traveled in backwards, sideways and down-up directions were not the same, the resulting $D^{*}_{AM}$ values were different. 

\begin{figure}\vspace{-0mm}
    \centering
    \resizebox{0.87\columnwidth}{!}{\rotatebox{0}{\includegraphics[trim=0cm 0cm 0cm 0cm, clip=true]{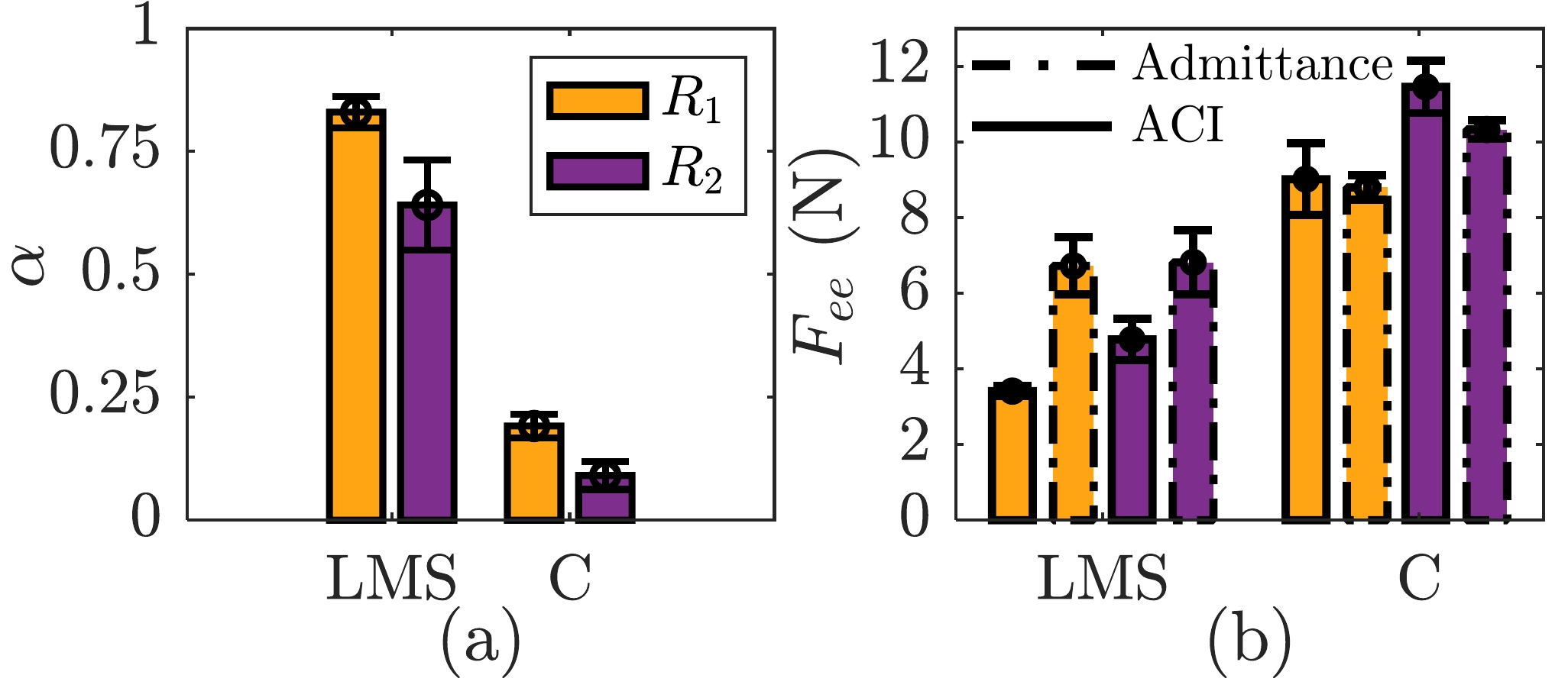}}}\vspace{-2mm}
     \caption{The means and standard errors of the adaptive index, $\alpha$, (a) and the measured force amplitudes (b) from the end-effector of the robots ($R_{1}$: yellow color, $R_{2}$: purple color) during the co-transportation of a bulky box carried by forklift moving straps (FMS), and a closet (C) directly grasp by the robots and human. The dashed and solid lines stand for the admittance controller and the ACI, respectively.}
    \vspace{-0.7cm}
    \label{fig:alfa_force}
\end{figure}

Fig.~\ref{fig:alfa_force} depicts the means and standard errors of the robots' adaptive indices, $\alpha$, and force amplitudes measured at their end-effectors, $F_{ee}$, for both experiments.
The resulting $\alpha$ values were much higher for the co-transportation of a bulky box with forklift moving straps (FMS) compared to the closet (C) directly grasped by the SoftHands (see Fig.~\ref{fig:alfa_force}(a)). 
As can be seen, when the connections between the human and the robots were rigid, the observed $\alpha$ values were close to zero. Here, the forces can be adequately transmitted through the jointly-held solid object (i.e., closet), and at the same time, $v_{adm}$ became more dominant w.r.t $v_{h}$ within the desired motion of the robots. 
On the other hand, the deformability brought to the system by the straps caused higher $\alpha$ values, as expected. 
Since $\alpha$ values approached to 1, the human kinematic information contributed more to the motion of the robots with respect to the less reliable haptic information.
The results in Fig.~\ref{fig:alfa_force}(b) show the measured mean end-effector forces of both robots were similar during the transportation of the closet.  
On the contrary, when the admittance controller was used instead of the ACI to successfully co-carry the bulky box with FMS throughout the desired path, the participants were required to convey higher force to their robot partners.

\vspace{-1mm}

\section{CONCLUSION}
\label{sec:conclusion}

In this study, we proposed an adaptive framework for human-multi-robot collaborative transportation of objects irrespective of their deformation characteristics. Our framework allows the human operator to collaborate with any number of robots when it is not feasible to manipulate the object due to its size and/or load with one partner. The performance of our framework during the co-transportation of objects with the help of the deformable straps shows its promising usability even in the cases with ungraspable ones that need to be handled. Future works will focus on adding an obstacle avoidance feature for the bases of mobile robots to make the framework more suitable for industrial environments.

\bibliographystyle{ieeetr}
\bibliography{biblio.bib}

\end{document}